\documentclass{article}

\usepackage[final, nonatbib]{neurips_2020}
\usepackage[square,numbers]{natbib}

\usepackage[utf8]{inputenc} 
\usepackage[T1]{fontenc}    
\usepackage{hyperref}       
\usepackage{url}            
\usepackage{booktabs}       
\usepackage{amsfonts}       
\usepackage{nicefrac}       
\usepackage{microtype}      
\usepackage[ruled,vlined]{algorithm2e}

\usepackage{amsmath,amsfonts,bm}

\def\eqref#1{equation~\ref{#1}}

\def\1{\bm{1}}

\DeclareMathAlphabet{\mathsfit}{\encodingdefault}{\sfdefault}{m}{sl}
\SetMathAlphabet{\mathsfit}{bold}{\encodingdefault}{\sfdefault}{bx}{n}

\def\gA{{\mathcal{A}}}

\def\gD{{\mathcal{D}}}

\def\gG{{\mathcal{G}}}

\def\gI{{\mathcal{I}}}

\def\gN{{\mathcal{N}}}
\def\gO{{\mathcal{O}}}

\def\gR{{\mathcal{R}}}
\def\gS{{\mathcal{S}}}
\def\gT{{\mathcal{T}}}
\def\gU{{\mathcal{U}}}

\def\gY{{\mathcal{Y}}}
\def\gZ{{\mathcal{Z}}}

\def\sP{{\mathbb{P}}}

\newcommand{\E}{\mathbb{E}}

\DeclareMathOperator*{\argmin}{arg\,min}

\usepackage{amsmath,amsfonts,bm}

\def\EX{{\mathbb{E}}}

\makeatletter
\newcommand{\subalign}[1]{%
  \vcenter{%
    \Let@ \restore@math@cr \default@tag
    \baselineskip\fontdimen10 \scriptfont\tw@
    \advance\baselineskip\fontdimen12 \scriptfont\tw@
    \lineskip\thr@@\fontdimen8 \scriptfont\thr@@
    \lineskiplimit\lineskip
    \ialign{\hfil$\m@th\scriptstyle##$&$\m@th\scriptstyle{}##$\hfil\crcr
      #1\crcr
    }%
  }%
}
\makeatother

\usepackage{appendix}      
\usepackage{graphicx} 
\usepackage{wrapfig}
\usepackage{amsmath}
\usepackage{tabularx}
\usepackage{multirow}

\usepackage{lipsum}
\usepackage{xcolor}

\usepackage{caption}
\usepackage{subcaption}

\usepackage{url}

\title{Reinforcement Learning for Robotic Manipulation using Simulated Locomotion Demonstrations}

\author{%
  Ozsel Kilinc \\
  WMG\\
  University of Warwick\\
  Coventry, UK CV4 7AL \\
  \texttt{ozsel.kilinc@warwick.ac.uk} \\
   \And
  Giovanni Montana \\
  WMG\\
  University of Warwick\\
  Coventry, UK CV4 7AL \\
  \texttt{g.montana@warwick.ac.uk} \\
}

\begin{document}

\maketitle

\begin{abstract}

Mastering robotic manipulation skills through reinforcement learning (RL) typically requires the design of shaped reward functions. Recent developments in this area have demonstrated that using sparse rewards, i.e. rewarding the agent only when the task has been successfully completed, can lead to better policies. However, state-action space exploration is more difficult in this case. Recent RL approaches to learning with sparse rewards have leveraged high-quality human demonstrations for the task, but these can be costly, time consuming or even impossible to obtain. In this paper, we propose a novel and effective approach that does not require human demonstrations. We observe that every robotic manipulation task could be seen as involving a locomotion task from the perspective of the object being manipulated, i.e. the object could learn how to reach a target state on its own. In order to exploit this idea, we introduce a framework whereby an object locomotion policy is initially obtained using a realistic physics simulator. This policy is then used to generate auxiliary rewards, called simulated locomotion demonstration rewards (SLDRs), which enable us to learn the robot manipulation policy. The proposed approach has been evaluated on 13 tasks of increasing complexity, and can achieve higher success rate and faster learning rates compared to alternative algorithms. SLDRs are especially beneficial for tasks like multi-object stacking and non-rigid object manipulation.
 
\end{abstract}

\section{Introduction}

Reinforcement Learning (RL) solves sequential decision-making problems by learning a policy that maximises expected rewards. Recently, with the aid of deep artificial neural network as function approximators, RL-trained agents have been able to autonomously master a number of complex tasks, most notably playing video games \cite{MnihKSRVBGRFOPB15} and board games \cite{SilverHMGSDSAPL16}. Robot manipulation has been extensively studied in RL, but is particularly challenging to master because it often involves multiple stages (e.g. stacking multiple blocks), high-dimensional state spaces (e.g. dexterous hand manipulation \cite{zhu2018dexterous, andrychowicz2018learning}) and complex dynamics (e.g.  manipulating non-rigid objects). Although promising performance has been reported on a wide range of tasks like grasping \cite{levine2016end, PopovHLHBVLTER17}, stacking \cite{NairMAZA18} and dexterous hand manipulation \cite{zhu2018dexterous, andrychowicz2018learning}, the learning algorithms usually require carefully-designed reward signals to learn good policies. For example, \cite{PopovHLHBVLTER17} propose a thoroughly weighted 5-term reward formula for learning to stack Lego blocks and \cite{GuHLL17} use a 3-term shaped reward to perform door-opening tasks with a robot arm. The requirement of hand-engineered, dense reward functions limits the applicability of RL in real-world robot manipulation to cases where task-specific knowledge can be captured.

The alternative to designing shaped rewards consists of learning with only sparse feedback signals, i.e. a non-zero rewards indicating the completion of a task. Using sparse rewards is more desirable in practise as it generalises to many tasks without the need for hand-engineering \cite{SilverHMGSDSAPL16,AndrychowiczCRS17,burda2018exploration}. On the other hand, learning with only sparse rewards is significantly more challenging since associating sequences of actions to non-zero rewards received only when a task has been successfully completed becomes more difficult. A number of existing approaches that address this problem have been proposed lately \cite{AndrychowiczCRS17,RiedmillerHLNDW18,MaxInfoExploration_Houthooft_16,burda2018large,pathak2017curiosity,burda2018exploration,savinov2018episodic,SOLAR_Zhang_19}; some of them report some success in completing manipulation tasks like object pushing \cite{AndrychowiczCRS17,SOLAR_Zhang_19}, pick-and-place \cite{AndrychowiczCRS17}, stacking two blocks \cite{RiedmillerHLNDW18,SOLAR_Zhang_19}, and target finding in a scene~\cite{pathak2017curiosity,savinov2018episodic}. Nevertheless, for more complex tasks such as stacking multiple blocks and manipulating non-rigid objects, there is scope for further improvement.

A particularly promising approach to facilitate learning has been to leverage human expertise through a number of manually generated examples demonstrating the robot actions required to complete a given task. When these demonstrations are available, they can be used by an agent in various ways, e.g. by attempting to generate a policy that mimics them \cite{RossGB11,ImitationLearningSelfDriving2_Bojarski_17,ImitationLearningSelfDriving1_Xu_17}, pre-learning a policy from them for further RL \cite{SilverHMGSDSAPL16,HesterVPLSPHQSO18}, as a mechanism to guide exploration \cite{NairMAZA18}, as data from which to infer a reward function ~\cite{NgR00,IRL_Abbeel_04,FinnLA16,HoE16}, and in combination with trajectories generated during RL \cite{VecerikHSWPPHRL17,LearningFromDemonstrationMistureSamples_Sasaki_19,LearningFromDemonstrationMistureSamples_Reddy_19}. Practically, however, human demonstrations are expensive to obtain, and their effectiveness ultimately depends on the competence of the demonstrators. Demonstrators with insufficient task-specific expertise could generate low-quality demonstrations resulting in sub-optimal policies. Although there is an existing body of work focusing on learning with imperfect demonstrations \cite{LearningFromImperfectDemonstration_Grollman_11,LearningFromImperfectDemonstration_Zheng_14,LearningFromImperfectDemonstration_Shiarlis_16,LearningFromImperfectDemonstration_Gao_18,LearningFromImperfectDemonstration_Brown_19,LearningFromDemonstrationMistureSamples_Choi_19}, these methods usually assume that either qualitative evaluation metrics are available~\cite{LearningFromImperfectDemonstration_Grollman_11,LearningFromImperfectDemonstration_Shiarlis_16,LearningFromImperfectDemonstration_Brown_19} or that a substantial volume of demonstrations can be collected~\cite{LearningFromImperfectDemonstration_Zheng_14,LearningFromImperfectDemonstration_Gao_18,LearningFromDemonstrationMistureSamples_Choi_19}.

\begin{figure*}[t]
	\begin{center}
		\centerline{\includegraphics[width=\linewidth,trim={0cm 6.8cm 0.cm 2.5cm},clip]{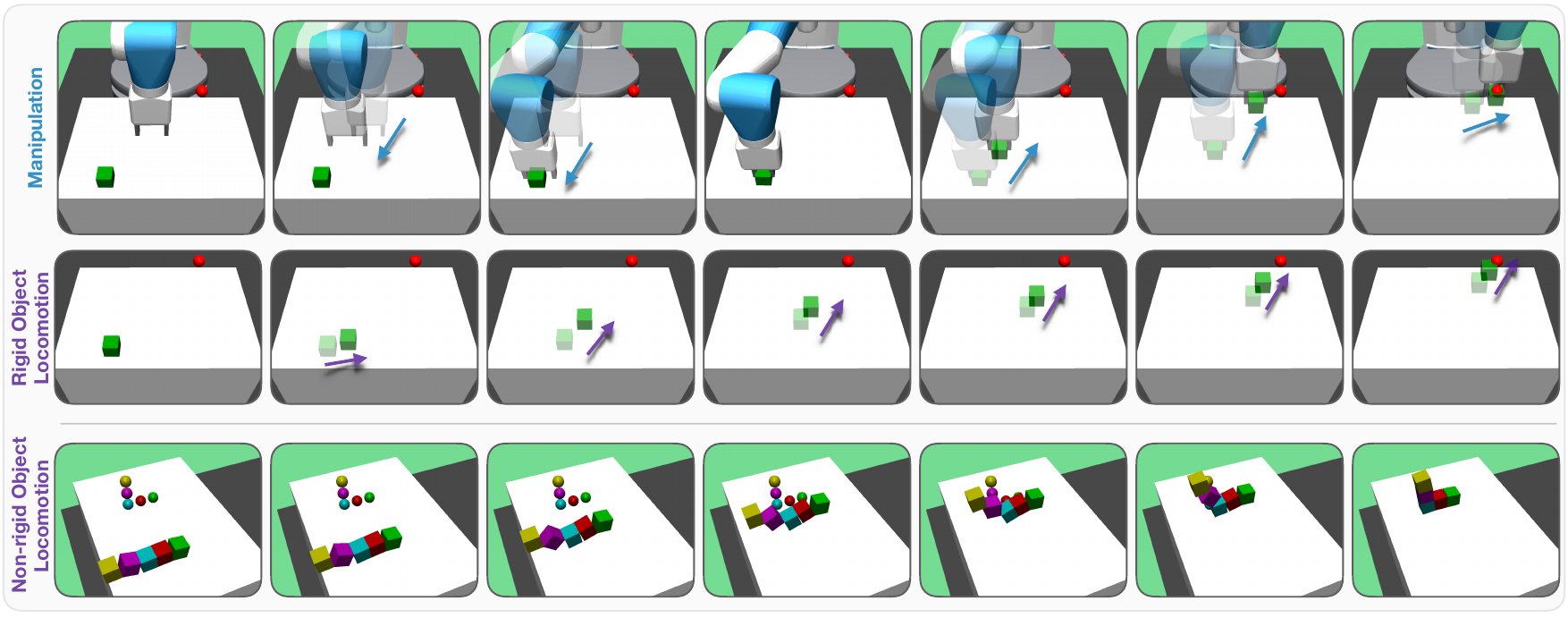}}
		\caption{An illustration of the proposed approach. Top row: a general robot manipulation task of pick-and-place, which requires the robot to pick up an object (green cube) and place it to a specified location (red sphere). Middle row: the corresponding auxiliary locomotion task requires the object to move to the target location. Bottom row: the auxiliary locomotion task corresponding to a pick-and-place task with a non-rigid object (not shown). Note that the auxiliary locomotion tasks usually have significantly simpler dynamics compared to the corresponding robot manipulation task, hence can be learnt efficiently through standard RL, even for very complex tasks. The learnt locomotion policy is used to inform the robot manipulation policy.}
		\label{fig:task_redefining}
	\end{center}
	\vskip -0.2in
\end{figure*}

In this paper, we propose a novel approach that allows complex robot manipulation tasks to be learnt with only sparse rewards.  In the tasks we consider, an object is manipulated by a robot so that, starting from a (random) initial position, it eventually reaches a goal position through a sequence of states in which its location and pose vary. For example, Figure \ref{fig:task_redefining} (top row) represents a pick-and-place task in which the object is being picked up by the two-finger gripper and moved from its initial state to a pre-defined target location (red sphere). Our key observation is that every robot manipulation implies an underlying object locomotion task that can be explicitly modelled as an independent task for the object itself to learn. Figure \ref{fig:task_redefining} (middle row) illustrates this idea for the pick-and-place task: the object, on its own, must learn to navigate from any given initial position until it reaches its target position. More complex manipulation tasks involving non-rigid objects can also be thought as inducing such object locomotion tasks; for instance, in Figure \ref{fig:task_redefining} (bottom row), a 5-tuple non-rigid object moves itself to the given target location and pose (see Figure \ref{fig:flex_obj_task} for the description of the non-rigid object).

Although in the real world it is impossible for objects to move on their own, learning such object locomotion policies can be achieved in a virtual environment through a realistic physics engine such as MuJoCo~\cite{todorov2012mujoco}, Gazebo~\cite{Gazebo_Koenig_2004} or Pybullet~\cite{Pybullet_Coumans_2017}.
In our experience, such policies are relatively straightforward to learn using only sparse rewards since the objects usually operate in simple state/action spaces and/or have simple dynamics. Once a locomotion policy has been learnt, we utilise it to produce a form of auxiliary rewards guiding the main manipulation policy. We name these auxiliary rewards ``Simulated Locomotion Demonstration Rewards'' (SLDRs). During the process of learning the robot manipulation policy, the proposed SLDRs encourage the robot to execute policies implying object trajectories that are similar to those obtained by the object locomotion policy. Although the SLDRs can only be learnt through a realistic simulator, this requirement does not restrict their applicability to real world problems, and the resulting manipulation policies can still be transferred to physical systems. To the best of our knowledge, this is the first time that object-level policies are trained in a physics simulator to enable robot manipulation learning driven by only sparse rewards.

In our implementation, all the policies are learnt using deep deterministic policy gradient (DDPG)~\cite{LillicrapHPHETS15}, which has been chosen due to its widely reported effectiveness in continuous control; however, most RL algorithms compatible with continuous actions could have been used within the proposed SLD framework. Our experimental results involve $13$ continuous control environments using the MuJoCo  physics engine~\cite{todorov2012mujoco} within the OpenAI Gym framework~\cite{brockman2016openai}. These environments cover a variety of robot manipulation tasks with increasing level of complexity, e.g. pushing, sliding and pick-and-place tasks with a Fetch robotic arm, in-hand object manipulation with a Shadow's dexterous hand, multi-object stacking, and non-rigid object manipulation. Overall, across all environments, we have found that our approach can achieve faster learning rate and higher success rate compared to baselines methods, especially in more challenging tasks such as stacking objects and manipulating non-rigid objects. Baselines are provided to represent existing approaches that use reward-shaping, curiosity-based auxiliary rewarding and auxiliary goal generation techniques.

The remainder of the paper is organised as follows. In Section~\ref{sec:relatedwork} we review the most related work, and in Section~\ref{sec:background} we provide some introductory background material regarding the RL modelling framework and algorithms we use. In Section~\ref{sec:method}, we develop the proposed methodology. In Section~\ref{sec:environments} we describe all the environments used for our experiments, and the experimental results are reported in Section~\ref{sec:experiments}. Finally, we conclude with a discussion and suggestions for further extensions in Section~\ref{sec:conclusion_discussion}.

\section{Related Work} \label{sec:relatedwork}

\textbf{Robotic Manipulation.} Robotics requires sequential decision making under uncertainty, and therefore it is a common application domain of machine learning approaches including RL \cite{kroemer2019review}. Recent advances in RL have focused on locomotion \cite{LillicrapHPHETS15, schulman2015trust, mnih2016asynchronous} and manipulation tasks \cite{fu2016one, GuHLL17}, which includes grasping \cite{levine2016end, PopovHLHBVLTER17}, stacking \cite{NairMAZA18} and dexterous hand manipulation \cite{zhu2018dexterous, andrychowicz2018learning}. These tasks are particularly challenging as they require continuous control over actions and the expected behaviours are hard to formulate through rewards. Due to the sample inefficiency problem of RL, most state-of-the-art approaches rely on simulated environments such as MuJoCo \cite{todorov2012mujoco} as training using physical systems would be significantly slower and costly. Predicting how objects behave under manipulation has also been well studied. For example, \cite{kopicki2009prediction, kopicki2011learning, belter2014kinematically} propose approaches to predict the motions of rigid objects under pushing actions with the aim of using these models to plan the robotic manipulation. Most recently, \cite{li2018learning} has proposed to learn a particle-based dynamics from data to handle complex interactions between rigid bodies, deformable objects and fluids. The focus of these studies has been to develop learnable simulators to replace traditional physics engines, whereas in this paper our aim is to learn object policies using the simulators. Although we employ a traditional physics engine for this paper, this could be replaced with learnable simulators in future work. \\
\textbf{Learning from Demonstrations.} 
A substantial body of work exists on how to leverage such demonstrations, when available, for reinforcement learning. Behaviour cloning (BC) methods approach sequential decision-making as a supervised learning problem  \cite{Pomerleau88, DuanASHSSAZ17,ImitationLearningSelfDriving2_Bojarski_17,ImitationLearningSelfDriving1_Xu_17}. Some BC methods include an expert demonstrator in the training loop to handle the mismatching between the demonstration data and the data encountered in the training procedure \cite{RossGB11, RatliffBS07}. Recent BC methods have also considered adversarial frameworks to improve the policy learning \cite{HoE16,ImitationLearningAdv_Wang_2018}. A different approach consists of inverse reinforcement learning, which seeks to infer a reward/cost function to guide the policy learning \cite{NgR00,IRL_Abbeel_04,FinnLA16}. Several methods have been developed to leverage demonstrations for robotic manipulation tasks with sparse rewards. For instance, \cite{VecerikHSWPPHRL17,NairMAZA18} jointly use demonstrations with trajectories collected during the RL process to guide the exploration, and \cite{HesterVPLSPHQSO18} use the demonstrations to pre-learn a policy, which is further fine-tuned in a following RL stage. Obtaining the training data requires specialised data capture setups such as teleoperation interfaces. In general, obtaining good quality demonstrations is an expensive process in terms of both human effort and equipment requirements. In contrast, the proposed method generates object-level demonstrations autonomously, and could potentially be used jointly with human-generated demonstrations when these are available.\\
\textbf{Goal Conditioned Policies and Auxiliary Goal Generation.} Goal-conditioned policies \cite{SchaulHGS15} that can generalise over multiple goals have been shown to be promising for robotic problems. For manipulation tasks with sparse rewards, several approaches have recently been proposed to automatically generate the auxiliary goals. For instance, \cite{SukhbaatarLKSSF18} used a self-play approach on reversible or resettable environments, \cite{FlorensaHGA18} employed adversarial training for robotic locomotion tasks, \cite{NairPDBLL18} proposed  variational autoencoders for visual robotics tasks, and \cite{AndrychowiczCRS17} introduced Hindsight Experience Replay (HER), which randomly draws synthetic goals from previously encountered states. HER in particular has been proved particularly effective, although the automatic goal generation can still be problematic on complex tasks involving multiple stages, e.g. stacking multiple objects, when used without demonstrations \cite{NairPDBLL18}. Some attempts have been made to form an explicit curriculum for such complex tasks; e.g. \cite{RiedmillerHLNDW18} manually define several semantically grounded sub-tasks each having its own individual reward. Methods such as this one requires significant human effort hence they cannot be readily  applied across different tasks. The proposed method in this paper uses goal-conditioned policies and adopts HER for auxiliary goal generation due to its effectiveness in robotic manipulation. However, it can be integrated with the other goal techniques in the literature. \\
\textbf{Auxiliary Rewards in RL.} Lately, increasing efforts have been made to design general auxiliary reward functions aimed at facilitating learning in environments with only sparse rewards. Many of these strategies involve a notion of curiosity \cite{schmidhuber1991curious}, which encourages agents to visit novel states that have not been seen in previous experience; for instance, \cite{pathak2017curiosity} formulate the auxiliary reward using the error in predicting the RL agent's actions by an inverse dynamics model,  \cite{MaxInfoExploration_Houthooft_16} encourage the agent to visit the states that result the largest information gain in system dynamics, \cite{burda2018exploration} construct the auxiliary reward based on the error in predicting the output of a fixed randomly initialised neural network, and \cite{savinov2018episodic} introduces the notion of state reachability. Despite the benefits introduced by these approaches, visiting unseen states may be less beneficial in robot manipulation tasks as exploring complex state spaces to find rewards is rather impractical \cite{AndrychowiczCRS17}. The proposed approach, on the other hand, produces the auxiliary rewards based on the underlying object locomotion; as such, it motivates the robot to mimic the optimal object locomotion rather than curiously exploring the continuous state space. 

\section{Background}\label{sec:background}
\subsection{Multi-goal RL for Robotic Manipulation}
We are concerned with solving a manipulation task: an object is presented to the robot, and has to be manipulated so as to reach a target position. In the tasks we consider, the target goal is specified by the object location and orientation, and the robot is rewarded only when it reaches its goal. We model the robot's sequential decision process as a Markov Decision Process (MDP) defined by a tuple, $M=\langle \gS, \gG, \gA, \gT, \gR, \gamma \rangle$, where $\gS$ is the set of states, $\gG$ is the set of goals, $\gA$ is the set of actions, $\gT$ is the state transition function, $\gR$ is the reward function and $\gamma \in \left[0, 1\right)$ is the discounting factor. At the beginning of an episode, the environment samples a goal $g \in \gG$. The state of the environment at time $t$ is denoted by $s_t \in \gS$ and includes both robot-related and object-related features. In a real system, these features are typically continuous variables obtained through sensors of the robot. The position of the object $o_t$ is one of the object-related features included in $s_t$ and can be obtained through a known mapping, i.e. $o_t=m_{\gS \rightarrow \gO}(s_t)$. A robot's action is controlled by a deterministic policy, i.e. $a_t=\mu_\theta(s_t,g): \gS \times \gG \rightarrow \gA$, parameterised by $\theta$. The environment moves to its next state through its state transition function, i.e. $s_{t+1}=\gT(s_t,a_t): \gS \times \gA \rightarrow \gS$, and provides an immediate and sparse reward $r_t$, defined as
\begin{equation}
    r_t = \gR(o_{t+1}, g) = 
    \begin{cases}
    0,      & \text{if } ||o_{t+1} - g||_2 \leq \epsilon \\
    -1,     & \text{otherwise}
\end{cases}
\end{equation}
where $\epsilon$ is a pre-defined threshold. Following its policy, the robot interacts with the environment until the episode terminates after $T$ steps. The interaction between the robot and the environment generates a trajectory, $\tau=(g,s_1,a_1,r_1,\hdots,s_T,a_T,r_T,s_{T+1})$. The ultimate learning objective is to find the optimal policy that maximises the expected sum of the discounted rewards over the time horizon $T$, i.e. 
\begin{equation}
\label{eq:rewards}
J(\mu_\theta) = \E_{\tau \sim  \sP(\tau|\mu_\theta)}[  R(\tau)=\sum_{i=1}^T{\gamma^{i-1}r_i}  ]
\end{equation}
where $\gamma$ is the discount factor. 

\subsection{Deep Deterministic Policy Gradient Algorithm}
\label{sec:background_rl_ddpg}

Policy Gradient (PG) algorithms update the policy parameters $\theta$ in the direction of $\nabla_\theta J(\mu_\theta)$ to maximise the expected return $J(\mu_\theta) = \E_{\tau \sim  \sP(\tau|\mu_\theta)}[R(\tau)]$. 
Deep Deterministic Policy Gradient (DDPG) \citep{LillicrapHPHETS15} integrates non-linear function approximators such as neural networks with Deterministic Policy Gradient (DPG) \citep{SilverLHDWR14} that uses deterministic policy functions. DDPG maintains a policy (actor) network $\mu_\theta(s_t,g)$ and an action-value (critic) network $Q^\mu(s_t,a_t,g)$. 

The actor $\mu_\theta(s_t,g)$ deterministically maps states to actions. The critic $Q^\mu(s_t,a_t,g)$ estimates the expected return when starting from $s_t$ by taking $a_t$, and then following $\mu_\theta$ in the future states until the termination of the episode, i.e. $Q^\mu(s_t,a_t,g) = \EX \big[\sum_{i=t}^T{\gamma^{i-t}r_i} \big| s_t,a_t,g,\mu_\theta \big]$. When interacting with the environment, DDPG assures the exploration by adding a noise to the deterministic policy output, i.e. $a_t = \mu_\theta(s_t,g) + \gN$. Experienced transitions during these interactions, i.e. $\langle g,s_t,a_t,r_t,s_{t+1}\rangle$, are stored in a replay buffer $\gD$. The actor and critic networks are updated using the transitions sampled from $\gD$. The critic parameters are learnt by minimising the following loss to satisfy the Bellman equation similarly to Q-learning \citep{WatkinsD92}:
\begin{equation}
    \label{eq:critic}
    \begin{split}
    \mathcal{L}(Q^\mu) = \EX_{g,s_t,a_t,r_t,s_{t+1} \sim \gD} \Big[(Q^\mu (s_t,a_t,g) - &y)^2 \Big]
    \end{split}
\end{equation}
where $y = r_t + \gamma Q^{\mu}(s_t,\mu(s_{t+1}),g)$. 
The actor parameters $\theta$ are updated using the following policy gradient:
\begin{equation}
    \label{eq:actor}
    \nabla_{\theta} J(\theta) = \EX_{g,s_t \sim \gD} \Big[\nabla_a Q^\mu(s_t,a,g)|_{a=\mu_\theta(s_t,g)} \nabla_{\theta}\mu_\theta(s_t,g)\Big]
\end{equation}
We adopt DDPG as the main training algorithm; however, the proposed idea can also be used with other off-policy approaches that work with continuous action domains.

\subsection{Hindsight Experience Replay}
\label{sec:background_rl_her}

Hindsight Experience Replay (HER) \cite{AndrychowiczCRS17} has been introduced to learn policies from sparse rewards, especially for robot manipulation tasks. The idea is to view the states achieved in an episode as pseudo goals (i.e. achieved goals) to facilitate learning even when the desired goal has not been achieved during the episode. Suppose we are given an observed trajectory, $\tau=(g,s_1,a_1,r_1,\hdots,s_T,a_T,r_T,s_{T+1})$. Since $o_t$ can be obtained from $s_t$ using a fixed and known mapping, the path that was followed by the object during the trajectory, i.e. $o_1,\hdots,o_{T+1}$, can be easily  extracted. HER samples a new goal from this path, i.e. $\Tilde{g} \sim \{o_1,\hdots,o_T\}$, and the rewards are recomputed with respect to $\Tilde{g}$, i.e. $\Tilde{r}_t = \gR(o_{t+1}, \Tilde{g})$. Using these rewards and $\Tilde{g}$, a new trajectory is created implicitly, i.e.  $\Tilde{\tau}=(\Tilde{g},s_1,a_1,\Tilde{r}_1,\hdots,s_T,a_T,\Tilde{r}_T,s_{T+1})$. These HER trajectories $\Tilde{\tau}$ are used to train the policy parameters together with the original trajectories.

\section{Methodology}
\label{sec:method}

Given a manipulation task, initially we introduce a corresponding auxiliary locomotion task for the object that is being manipulated, i.e. the object is assumed to be the decision-making agent. This auxiliary problem is usually significantly easier to learn compared to the original task. After learning the object locomotion policy, we use it on a reward-generating mechanism for the robot when learning the original manipulation task. In this section, we explain the steps involved in our proposed procedure, i.e. (a) how the object locomotion policies are learned, (b) how the proposed reward function is defined, and (c) how these auxiliary rewards are leveraged for robotic manipulation.

\subsection{Object Locomotion Policies}
\label{sec:object_locomotion}

The object involved in the manipulation task is initially modelled as an agent capable of independent decision making abilities, and its decision process is modelled by a separate MDP defined by a tuple $L=\langle \gZ, \gG, \gU, \gY, \gR, \gamma \rangle$. Here, $\gZ$ is the set of states, $\gG$ is the set of goals, $\gU$ is the set of actions, $\gY$ is the state transition function, $\gR$ is the reward function and $\gamma \in \left[0, 1\right)$ is the discounting factor. The same goal space, $\gG$, is used as in $M$, and $z_t \in \gZ$ is a reduced version of $s_t$ that only involves object-related features including the position of the object, i.e. $o_t \subset z_t$. The object's action space explicitly controls the pose of the object, and these actions are controlled by a deterministic policy, i.e. $u_t=\nu_\theta(z_t,g): \gZ \times \gG \rightarrow \gU$. The state transition is defined on a different space, i.e. $\gY: \gZ \times \gU \rightarrow \gZ$; however, the same sparse reward function is used here as before. Figure \ref{fig:block_obj_rob}a illustrates the training procedure used in this context and based on DDPG with HER. The optimal object policy $\nu_\theta$  maximises the expected return $J\left(\nu_\theta \right) = \E_{g,z_t \sim  \gD_L}[\sum_{i=1}^T{\gamma^{i-1}r_i}]$ where $\gD_L$ denotes the replay buffer containing the trajectories, indicated by $\eta$, obtained by $\nu_\theta$ throughout training.

\subsection{Robotic Manipulation with Simulated Locomotion Demonstration Rewards (SLDR)}
\label{sec:robot_policy_learning}

On the original manipulation task $M$, the robot receives the current environmental state and the desired goal and then decides how to act according to its policy $\mu_\theta$. Whenever the object is moved from one position to another, the observed object locomotion is a consequence of robot's actions. More concretely, the observed object action on $M$ (hereafter denoted by $w_t$) is a function of the robot policy $\mu_\theta$. The relation between $w_t$ and $\mu_\theta$ depends on the environmental dynamics whose close-form model is unknown. We use $f:\gA \rightarrow \gU$ to denote this unknown relation, i.e. $w_t=f(\mu_\theta(s_t,g))$.

The key steps of the proposed approach are as follows: as we had initially learnt an object locomotion policy on $L$, first we use it to enquire the optimal object action for the current state and goal, i.e. $u_t=\nu_\theta(z_t,g)$. Then, we update $\mu_\theta$ in order to make $w_t$ get closer to $u_t$. This learning objective can be written as follows:
\begin{equation}
    \label{eq:dpl_ours_modified}
    \argmin_{\mu_\theta} \mathbb{E}_{s_t}\big|\big|w_t-u_t\big|\big|^2_2 =
    \argmin_{\mu_\theta} \mathbb{E}_{s_t \sim P(s_t|\mu_\theta)}\Big|\Big|f\big(\mu_\theta(s_t,g)\big)-\nu_\theta(z_t,g)\Big|\Big|^2_2
\end{equation}

Given that the the environment dynamics is unknown, we replace $f$ in Eq. \ref{eq:dpl_ours_modified} with a parameterised model to approximate $w_t$. Estimating $w_t$ from robot actions is not straight-forward as it requires keeping track of all previous actions, i.e. $a_{1:t}$, and the initial state. Instead, we propose to estimate $w_t$ by evaluating the transition from the current state to the next. Specifically, we substitute $f$ with a parameterised inverse dynamic model, i.e. $\gI_\phi:\gZ \times \gZ \rightarrow \gU$, that we train to estimate the output of $\nu_\theta(z_t,g)$ from $z_t$ and $z_{t+1}$, i.e. $\nu_\theta(z_t,g) \approx \gI_\phi(z_t, z_{t+1})$. We learn the parameters of $\gI_\phi$ on the object locomotion task $L$ (see Section~\ref{sec:learning_algorithm} and Algorithm \ref{algo:obj_training} for training details), and then employ the trained model on the manipulation task $M$ to approximate $w_t$. Substituting $\gI_\phi$ into Eq.~\ref{eq:dpl_ours_modified} leads to the following optimisation problem:

\begin{equation}
    \label{eq:dpl_ours_appx_modified}
    \argmin_{\mu_\theta} \mathbb{E}_{s_t}\big|\big|w_t-u_t\big|\big|^2_2 \approx
    \argmin_{\mu_\theta} \mathbb{E}_{s_t \sim P(s_t|\mu_\theta)}\Big|\Big|\gI_\phi(z_t, z_{t+1})-\nu_\theta(z_t,g)\Big|\Big|^2_2
\end{equation}
On $M$, $z_{t+1}$ is a function of $\gT(s_t, \mu_\theta(s_t,g))$. In our setting, the close-form of the state transition function $\gT$ is unknown, instead $\gT$ can only be sampled. Also, pursuing a model-free approach, we do not aim to learn a model for $\gT$. Therefore, minimising Eq.~\ref{eq:dpl_ours_appx_modified} through gradient-based methods is not an option for our setting as this would require differentiation through $\gT$. Instead, we propose to formalise this objective as a reward to be maximised through a standard model-free RL approach. The first obvious candidate for this reward notion can be written as follows: 
\begin{equation}
    \label{eq:r_aux_1}
    q_t = - \big|\big|w_t-u_t\big|\big|^2_2 \approx - \Big|\Big|\gI_\phi(z_t, z_{t+1})-\nu_\theta(z_t,g)\Big|\Big|^2_2
\end{equation}
Practically, however, the above reward is sensitive to the scales of $\gI_\phi(z_t, z_{t+1})$ and $\nu_\theta(z_t,g)$, and therefore it may require an additional normalisation term. Even with a normalisation term, the scale of the rewards would shift throughout the training depending on the exploration and the sampling. In order to deal with this issue, we propose another reward notion adopting $Q^\nu$, i.e. the action-value function that had been learnt for $\nu_\theta$ on object locomotion task $L$ (see Section \ref{sec:learning_algorithm} and Algorithm \ref{algo:obj_training} for training details). The proposed reward notion is written as follows:
\begin{equation}
    \label{eq:r_aux_2}
    \begin{aligned}
    q_t^\text{\smaller{SLDR}} &= Q^\nu\big(z_t, w_t, g\big) - Q^\nu\big(z_t, u_t, g\big)\\
        &\approx  Q^\nu\big(z_t, \gI_\phi(z_t, z_{t+1}), g\big) - Q^\nu\big(z_t, \nu_\theta(z_t,g), g)\big)
    \end{aligned}
\end{equation}
We refer to Eq.~\ref{eq:r_aux_2} as the Simulated Locomotion Demonstrations Rewards (SLDR). Rather than comparing $w_t$ and $u_t$ directly with each other as in Eq.~\ref{eq:r_aux_1}, the SLDR compares their action-values using $Q^\nu$. Being learnt on $L$ using sparse rewards, $Q^\nu$ is well-bounded \cite{RLIntroBook2nd_Sutton_2018}, and $q_t^\text{\smaller{SLDR}}$ produced adopting $Q^\nu$ does not require a normalisation term. 

Note that, by definition, $Q^\nu(z_t,u,g)$ gives the expected return for any object locomotion action $u \in \gU$, when it is taken at the current state $z_t$ and then $\nu_\theta$ is followed for the future states. Since $\nu_\theta$ had been learnt through standard RL to maximise the sparse rewards, it is the optimal object locomotion policy, and therefore $Q^\nu(z_t, \nu_\theta(z_t,g), g))$ gives the maximum expected return. Accordingly, $q_t^\text{\smaller{SLDR}}$ can be viewed as the advantage of $w_t$ with respect to $\nu_\theta(z_t,g)$ in terms of the action-values, and is expected to be non-positive. Maximising this term encourages the robot to induce similar object actions compared to the optimal ones according to $\nu_\theta$.

\begin{algorithm}[tb]
	\begin{smaller}
		\DontPrintSemicolon
		\SetKwInOut{Given}{Given}
		\SetKwInOut{Initialise}{Initialise}
		\SetKwInOut{Return}{Return}
		
		\Given{Locomotion MDP $L=\langle \gZ, \gG, \gU, \gY, \gR, \gamma \rangle$, \\
		       Neural networks $\nu_\theta$, $Q^\nu$ and $\gI_\phi$ \\
		       A random process $\gN_L$ for exploration\\
		       Fixed and known mapping function $m_{\gZ \rightarrow \gO}:\gZ \rightarrow \gO$
		       }
		\Initialise{Parameters of $\nu_\theta$, $Q^\nu$ and $\gI_\phi$ \\
		            Experience replay buffer $\gD_L$\\
		            }
        \For{$i_{\textit{episode}} = 1$ \textbf{to} $N_{\textit{episode}}$}{
        \For{$i_{\textit{rollout}} = 1$ \textbf{to} $N_{\textit{rollout}}$}{
		    Receive initial state $z_1$ and $g$, $o_1=m_{\gZ \rightarrow \gO}(z_1)$\\
		    \For{$t = 1, T$}{
		        Sample an object action: $u_t = \nu_\theta(z_t,g)+\gN_L$ \\
		        Execute the action: $z_{t+1} = \gY(z_t,a_t)$, $r_t = \gR(o_{t+1},g)$\\
		        Store $(g,z_t,u_t,r_t,z_{t+1})$ in $\gD_L$ \\
		    }
		    Generate HER samples and store in $\gD_L$ \\ 
		}
		    \For{$i_{\textit{update}} \gets 1$ \textbf{to} $N_{\textit{update}}$}{
		      Get a random mini-batch of samples from $\gD_L$\\
		      Update $Q^\nu$ minimising the loss in Eq.(\ref{eq:critic})\\
		      Update $\nu_\theta$ using the gradient in Eq.(\ref{eq:actor})\\
		      Update $\gI_\phi$ minimising the loss in Eq.(\ref{eq:train_inverse})\\
		      }
		}
	    \Return{$\nu_\theta$, $Q^\nu$ and $\gI_\phi$}
		\caption{Learning locomotion policy and inverse dynamic}\label{algo:obj_training}
	\end{smaller}
\end{algorithm}

\subsection{Learning Algorithms} \label{sec:learning_algorithm}

\begin{figure}[t]
	\begin{center}
		\centerline{\includegraphics[width=\linewidth,trim={0.8cm 11cm 0.5cm 2.5cm},clip]{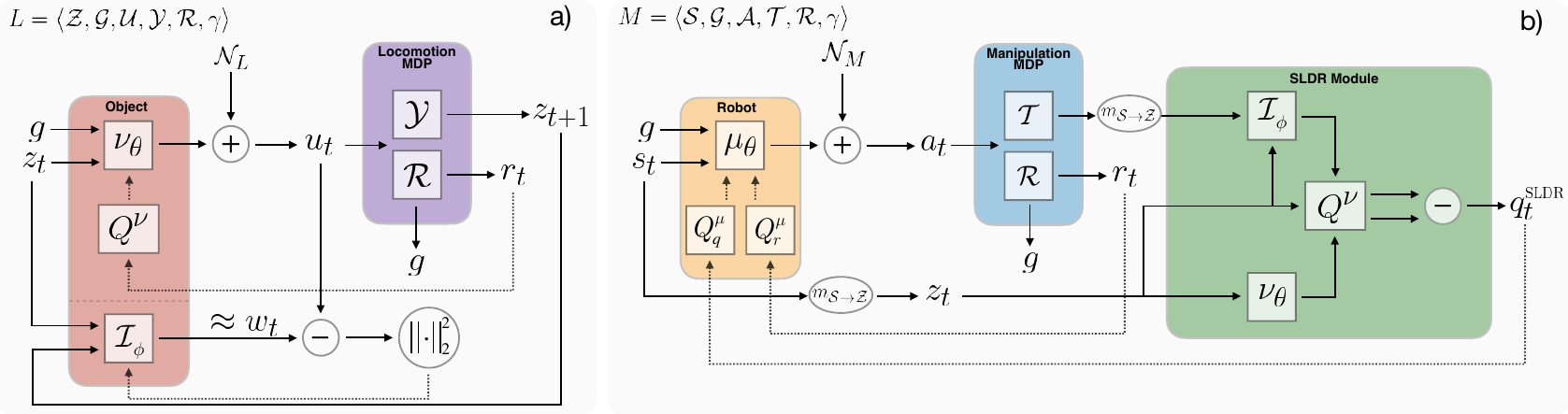}}
        \caption{Block diagram for the proposed SLD algorithm. The solid lines represent the forward pass of the model. The dashed lines represent the rewards/losses as feedback signals for model learning.}
		\label{fig:block_obj_rob}
	\end{center}
\end{figure}

In this subsection, we detail the learning algorithms for the object locomotion and the robotic manipulation policies. Figure \ref{fig:block_obj_rob} shows the block diagrams of the learning procedures.

{\bf Object locomotion policy}. We learn the object locomotion policy only using the environmental sparse rewards as described in Algorithm \ref{algo:obj_training}. We adopt DDPG (Section \ref{sec:background_rl_ddpg}) as the training framework together with HER (Section \ref{sec:background_rl_her}) to generate auxiliary transition samples to deal with the exploration difficulty caused by the sparse rewards. $Q^\nu$ is updated to minimise Eq. \ref{eq:critic}, and $\nu_\theta$ is optimised using the gradient in Eq.~\ref{eq:actor}. Concurrently, we learn $\gI_\phi$ using the trajectories generated during the policy learning process by minimising the following objective function:
\begin{equation}
\label{eq:train_inverse}
    \argmin_{\phi} \EX_{z_t,u_t, z_{t+1} \sim \gD_L} \Big|\Big|\gI_\phi(z_t, z_{t+1}) - u_t \Big|\Big|^2_2
\end{equation}
where $\gD_L$ is an experience replay buffer.

{\bf Robotic manipulation policy}. Similarly, we learn the robotic manipulation policy adopting DDPG with HER as described in Algorithm \ref{algo:bot_training}. Using the optimisation objective given in Eq. \ref{eq:critic}, we learn two action-value functions: $Q^\mu_r$ for the environmental sparse rewards $r_t$, and $Q^\mu_q$ for the proposed SLDR $q_t^\text{\smaller{SLDR}}$. Accordingly, $\mu_\theta$ is updated following the gradient below that uses both action-value functions:
\begin{equation}
\label{eq:policy_bot_modify}
    \nabla_{\theta} J(\theta) = \EX_{g,s_t \sim \gD_M} \Big[ \nabla_a\big(Q^{\mu}_{r}(s_t,a,g) + Q^{\mu}_{q}(s_t,a,g)\big)\big|_{a=\mu(s_t,g)} \nabla_\theta \mu_\theta(s_t,g)\Big]
\end{equation}
where $\gD_M$ is an experience replay buffer. Some tasks may include $N>1$ objects, e.g. stacking. The proposed method is able to handle these tasks by using individual SLDR for each object and learning individual $Q^\mu_{q_i}$ for each one of them. Then, the gradient required to update $\mu_\theta$ is:
\begin{equation}
\label{eq:policy_bot_multi_modify}
    \nabla_{\theta} J(\theta) = \EX_{g,s_t \sim \gD_M} \Big[ \nabla_a\big(Q^{\mu}_{r}(s_t,a,g) + \sum Q^{\mu}_{q_i}(s_t,a,g)\big)\big|_{a=\mu(s_t,g)} \nabla_\theta \mu_\theta(s_t,g)\Big]
\end{equation}

\begin{algorithm}[tb]
	\begin{smaller}
		\DontPrintSemicolon
		\SetKwInOut{Given}{Given}
		\SetKwInOut{Initialise}{Initialise}
		\SetKwInOut{Return}{Return}
		
		\Given{Manipulation MDP $M=\langle \gS, \gA, \gT, \gR, \gamma \rangle$, \\
		       Learnt object-related components $\nu_\theta$, $Q^\nu$ and $\gI_\phi$ \\
		       Neural networks $\mu_\theta$, $Q^\mu_r$ and $Q^\mu_q$\\
		       A random process $\gN_M$ for exploration \\
		       Fixed and known mapping function $m_{\gS \rightarrow \gO}:\gS \rightarrow \gO$ \\
		       Fixed and known mapping function $m_{\gS \rightarrow \gZ}:\gS \rightarrow \gZ$ \\
		       }
		\Initialise{Parameters of $\mu_\theta$, $Q^\mu_r$ and $Q^\mu_q$ \\
		            Experience replay buffer $\gD_M$\\
		            }
        \For{$i_{\textit{episode}} = 1$ \textbf{to} $N_{\textit{episode}}$}{
        \For{$i_{\textit{rollout}} = 1$ \textbf{to} $N_{\textit{rollout}}$}{
		    Receive initial state $s_1$ and $g$, $o_1=m_{\gS \rightarrow \gO}(s_1)$\\
		    \For{$t = 1, T$}{
		        Sample a robot action: $a_t = \mu_\theta(s_t,g)+\gN_M$ \\
		        Execute the action: $s_{t+1} = \gT(s_t,a_t)$, $r_t = \gR(o_{t+1},g)$\\
		        Obtain SLD reward: \\
		        $z_t=m_{\gS \rightarrow \gZ}(s_t)$ and $z_{t+1}=m_{\gS \rightarrow \gZ}(s_{t+1})$\\
		        $q_t^\text{\smaller{SLDR}} = Q^\nu\big(z_t, \gI_\phi(z_t, z_{t+1}), g) - Q^\nu\big(z_t, \nu_\theta(z_t,g),g\big)$ \\
		        Store $(g, s_t,a_t,r_t,q_t,s_{t+1})$ in $\gD_M$ \\
		    }
		    Generate HER samples and store in $\gD_M$ \\ 
		}
		    \For{$i_{\textit{update}} \gets 1$ \textbf{to} $N_{\textit{update}}$}{
		      Get a random mini-batch of samples from $\gD_M$\\
		      Update $Q^\mu_r$ minimising the loss in Eq.(\ref{eq:critic}) for $r_t$\\
		      Update $Q^\mu_q$ minimising the loss in Eq.(\ref{eq:critic}) for $q_t^\text{\smaller{SLDR}}$\\
		      Update $\mu_\theta$ using the gradient in Eq.(\ref{eq:policy_bot_modify})\\
		      }
		}
	    \Return{$\mu$}
		\caption{Learning manipulation policy}\label{algo:bot_training}
	\end{smaller}
\end{algorithm}

\section{Environments}
\label{sec:environments}

We have evaluated the SLD method on 13 simulated MuJoCo \cite{todorov2012mujoco} environments using two different robot configurations: 7-DoF Fetch robotic arm with a two-finger parallel gripper and 24-DoF Shadow's Dexterous Hand. The tasks we have chosen to evaluate include single rigid object manipulation, multiple rigid object stacking and non-rigid object manipulation. Overall, we have used 9 MuJoCo environments (3 with Fetch robot arm and 6 with Shadow's hand) for single rigid object tasks. Furthermore, we have included additional environments for multiple   object stacking and non-rigid object manipulation using the Fetch robot arm. In all environments the rewards are sparse.

\textbf{Fetch Arm Single Object Environments}. These are the same \textit{Push}, \textit{Slide} and \textit{PickAndPlace} tasks introduced in \cite{PlappertHER2}. In each episode, a desired 3D position (i.e. the target) of the object is randomly generated. The reward is zero if the object is within 5cm range to the target, otherwise $-1$. The robot actions are 4-dimensional: 3D for the desired arm movement in Cartesian coordinates and 1D to control the opening of the gripper. In pushing and sliding, the gripper is locked to prevent grasping. The observations include the positions and linear velocities of the robot arm and the gripper, the object's position, rotation, angular velocity, the object's relative position and linear velocity to the gripper, and the target coordinate. An episode terminates after $50$ time-steps. 

\textbf{Shadow's Hand Single Object Environments}. These include the tasks first introduced in~\cite{PlappertHER2}, i.e. \textit{Egg}, \textit{Block}, \textit{Pen} manipulation. In these tasks, the object (a block, an egg-shaped object, or a pen) is placed on the palm of the robot hand; the robot hand is required to manipulate the object to reach a target pose. The target pose is 7D describing the 3D position together with 4D quaternion orientation, and is randomly generated in each episode.  The reward is $0$ if the object is within some task-specific range to the target, otherwise $-1$. As in \cite{PlappertHER2}, each task has two variants: \textit{Full} and \textit{Rotate}. In the \textit{Full} variant, the object's whole 7D pose is required to meet the given target pose. In the \textit{Rotate} variants, the 3D object position is ignored and only the 4D object rotation is expected to the satisfy the desired target. Robot actions are 20-dimensional controlling the absolute positions of all non-coupled joints of the hand. The observations include the positions and velocities of all 24 joints of the robot hand, the object's position and rotation, the object's linear and angular velocities, and the target pose. An episode terminates after $100$ time-steps. 

\textbf{Fetch Arm Multiple Object Stacking Environments}. The stacking task is built upon the \textit{PickAndPlace} task. We consider 2- and 3-object stacking tasks. For $N$-object stacking task, the target has $3N$ dimensions describing the desired  positions of all $N$ objects in 3D. Following~\cite{NairMAZA18}, we start these tasks with the first object placed at its desired target. The robot needs to perform $N-1$ pick-and-place actions without displacing the first object. The reward is zero if all objects are within $5$cm range to their designated targets, otherwise the reward is assigned a value of $-1$. The robot actions and observations are similar to those in the \textit{PickAndPlace} task. The episode length is 50 time-steps for 2-object stacking and 100 for 3-object.

\begin{figure}[t]
	\begin{center}
		\centerline{\includegraphics[width=8cm,trim={0.5cm 12.2cm 18.3cm 2.9cm},clip]{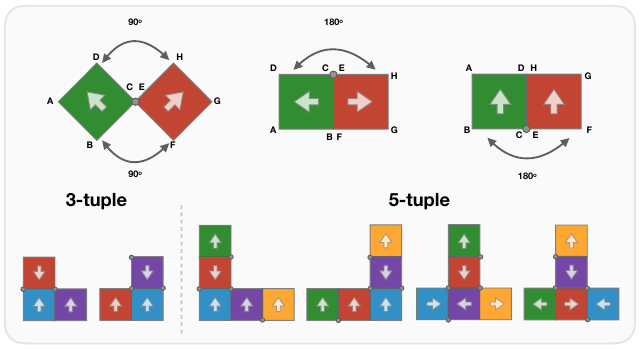}}
		\caption{An illustration of the non-rigid objects used in the experiments. Top row: a hinge joint (shown as grey circles) between two neighbouring blocks allows one rotational DoF along their coincident edges up to $180^o$. Bottom row: each variant has some predefined target poses (2 options for \textit{3-tuple} and 4 for \textit{5-tuple}).
		}
		\label{fig:flex_obj_task}
	\end{center}
	\vskip -0.3in
\end{figure}

\textbf{Fetch Arm Non-rigid Object Environments}. We build non-rigid object manipulation tasks based on the \textit{PickAndPlace} task. Instead of using the original rigid block, we have created a non-rigid object by hinging some blocks side-by-side along their edges as shown in Figure \ref{fig:flex_obj_task}. A hinge joint is placed between two neighbouring blocks, allowing one rotational degree of freedom (DoF) along their coincident edges up to $180^o$. We introduce two different variants: \textit{3-tuple} and \textit{5-tuple}. For the \textit{$N$-tuple} task, $N$ cubical blocks are connected with $N-1$ hinge joints creating $N-1$ internal DoF. The target pose has $3N$-dimension describing the desired 3D positions of all $N$ blocks, which are selected uniformly in each episode from a set of predefined target poses (see Figure \ref{fig:flex_obj_task}). The robot is required to manipulate the object to match the target pose. The reward is zero when all the $N$ blocks are within a $2$cm range to their corresponding targets, otherwise $-1$. Robot actions and observations are similar to those in the \textit{PickAndPlace} tasks, excepting that the observations include the position, rotation, angular velocity, relative position and linear velocity to the gripper for each block. The episode length is $50$ time-steps for both \textit{3-tuple} and \textit{5-tuple}.

\textbf{Object Locomotion Environments.} For each robotic manipulation task described above, we use an object locomotion task where we first learn $\nu_\theta$, $Q^\nu$ and $\gI_\phi$. Here, we detail the observation and action space differences between object locomotion and robotic manipulation tasks.

For any task, the object's observation is a subset of the robot's observation, i.e. $z_t \subset s_t$, and only includes object-related features while excluding those related to the robot. More concretely, for the environments with the Fetch arm, the object's observations include the object's position, rotation, angular velocity, the object's relative position and linear velocity to the target, and the target location. For the environments with the Shadow's hand, the object observations include the object's position and rotation, the object's linear and angular velocities, and the target pose. We define the object action as the desired relative change in the 7D object pose (3D position and 4D quaternion orientation) between two consecutive time-steps. This leads to 7D action spaces. Specifically for non-rigid objects, we define the object action as the desired relative change in the poses of the blocks at two ends. This leads to 14D action spaces. The rewards are the same as those in each robot manipulation task.

It is worth noting that, in the \textit{Full} variants of Shadow's hand environments, we consider the object translation and rotation as two individual locomotion tasks, and we learn separate locomotion policies and Q-functions for each task. We find that the above strategy encourages the manipulation policy to perform translation and rotation simultaneously. Although object translation and rotation could be executed within a single task, we have empirically found that the resulting manipulation policies tend to prioritise one behaviour versus the other (e.g. they tend to rotate the object first, then translate it) and generally achieves a lower performance. 

\section{Experiments} \label{sec:experiments}

\subsection{Implementation and Training Process}

Three-layer neural networks with ReLU activations was used to approximate all policies, action-value functions and inverse dynamics models. The Adam optimiser \cite{KingmaB14} was employed to train all the neural networks. During the training of locomotion policies, the robot was considered as a non-learning component in the scene and its actions were not restricted to prevent any potential collision with the objects. We could have different choices for the actions of the robot. For example, we could let the robot move randomly or perform any arbitrary fixed action (e.g. a robot arm moving upwards with constant velocity until it reaches to the maximum height and then staying there). In preliminary experiments, we assessed whether this choice bears any effect on final performance, and concluded that no particular setting had clear advantages. For learning locomotion and manipulation policies, most of the hyperparameters suggested in the original HER implementation \cite{PlappertHER2} were retained with only a couple of exceptions for locomotion policies only: to facilitate exploration, with probability $0.2$ ($0.3$ in \cite{PlappertHER2}) a random action was drawn from a uniform distribution, otherwise we retained the current action, and added Gaussian noise with zero mean and $0.05$ ($0.2$ in \cite{PlappertHER2}) standard deviation. For locomotion policies, in all Shadow's hand environments and \textit{5-tuple}, we train the objects over 50 epochs. In the remaining environments, we stop the training after 20 epochs. When training the main manipulation policies, the number of epochs varies across tasks. For both locomotion and manipulation policies, each epoch includes $50$ cycles, and each cycle includes $38$ rollouts generated in parallel through $38$ MPI workers using CPU cores. This leads to $38 \times 50=1900$ full episodes per epoch. For each epoch, the parameters are updated $40$ times using a batch size of $4864$ on a GPU core. We normalise the observations to have zero mean and unit standard deviation as input of neural networks. We update mean and standard deviations of the observations using running estimation on the data in each rollout. We clip $q_t^\text{\smaller{SLDR}}$ to the same range with the environmental sparse rewards, i.e. $\left[-1, 0\right]$.

Our algorithm has been implemented in PyTorch\footnote{\url{https://pytorch.org/}}. All the environments are based on OpenAI Gym. The corresponding source code, the environments, and illustrative videos for selected tasks have been made publicly available.\footnote{{Source code: } \url{https://github.com/WMGDataScience/sldr.git}}\footnote{{Environments: } \url{https://github.com/WMGDataScience/gym_wmgds.git}}\footnote{{Supplementary videos: }\url{https://youtu.be/jubZ0dPVl2M}}

\subsection{Comparison and Performance Evaluation}
We include the following methods for comparisons:
\begin{itemize}
\item\textbf{DDPG-Sparse:} Refers to DDPG \cite{LillicrapHPHETS15} using sparse rewards.
\item\textbf{HER-Sparse:} Refers to DDPG with HER \cite{AndrychowiczCRS17} using sparse rewards.
\item\textbf{CHER-Sparse:} Refers to DDPG with Curriculum-guided Hindsight Experience Replay \cite{fang2019curriculum} using sparse rewards.
\item\textbf{HER-Dense:} Refers to DDPG with HER, using dense distance-based rewards.
\item\textbf{DDPG-Sparse+SLDR:} Refers to DDPG using sparse environmental rewards and SLDR proposed in this paper.
\item\textbf{HER-Sparse+RNDR:} Refers to DDPG with HER, using sparse environmental rewards and random network distillation-based auxiliary rewards (RNDR) \cite{burda2018exploration}.
\item\textbf{HER-Sparse+SLDR:} Refers to DDPG with HER, using sparse environmental rewards and SLDR.
\end{itemize}
We use DDPG-Sparse, HER-Sparse and HER-Dense as baselines. HER-Sparse+RNDR is a representative method constructing auxiliary rewards to facilitate policy learning. CHER-Sparse replaces the random selection mechanism of HER with an adaptive one that considers the proximity to true goals. DDPG-Sparse+SLDR and HER-Sparse+SLDR represents the proposed approach using SLDR with different methods for policy learning.

Following \cite{PlappertHER2}, we evaluate the performance after each training epoch by performing 10 deterministic test rollouts for each one of the 38 MPI workers. Then we compute the test success rate by averaging across the 380 test rollouts. For all comparison methods, we evaluate the performance with 5 different random seeds and report the median test success rate with the interquartile range. In all environments, we also keep the models with the highest test success rate for different methods and compare their performance. 

\subsection{Single Rigid Object Environments}

\begin{figure}
\begin{center}
\begin{subfigure}[b]{0.49\columnwidth}
    \centering
    \includegraphics[width=\columnwidth,trim={0.1cm 0.2cm 0.3cm 0.2cm},clip]{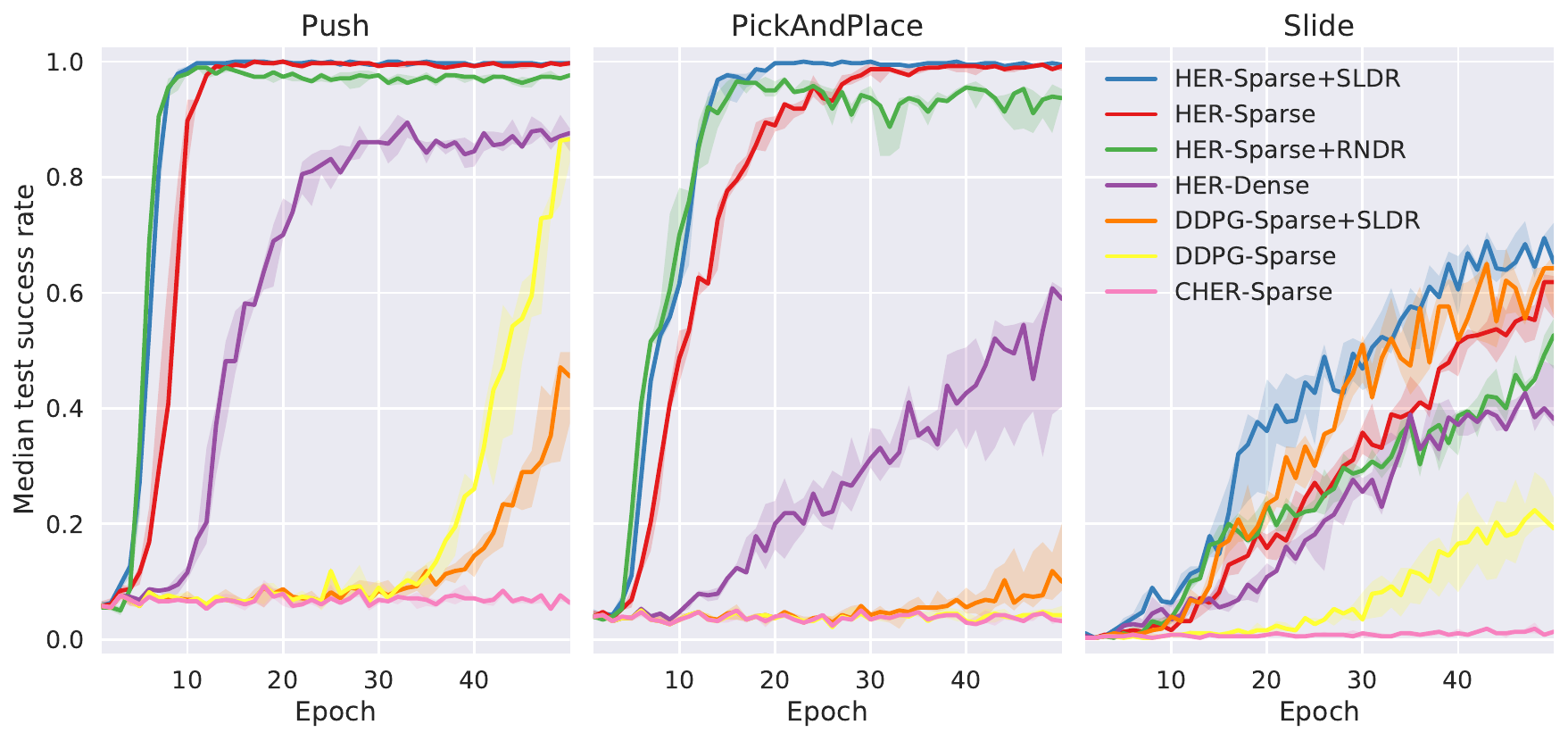}
    \caption{Fetch arm single object.}
    \label{fig:learning_fetch}
\end{subfigure}
\hfill
\begin{subfigure}[b]{0.49\columnwidth}
    \centering
    \includegraphics[width=\columnwidth,trim={0.1cm 0.2cm 0.3cm 0.2cm},clip]{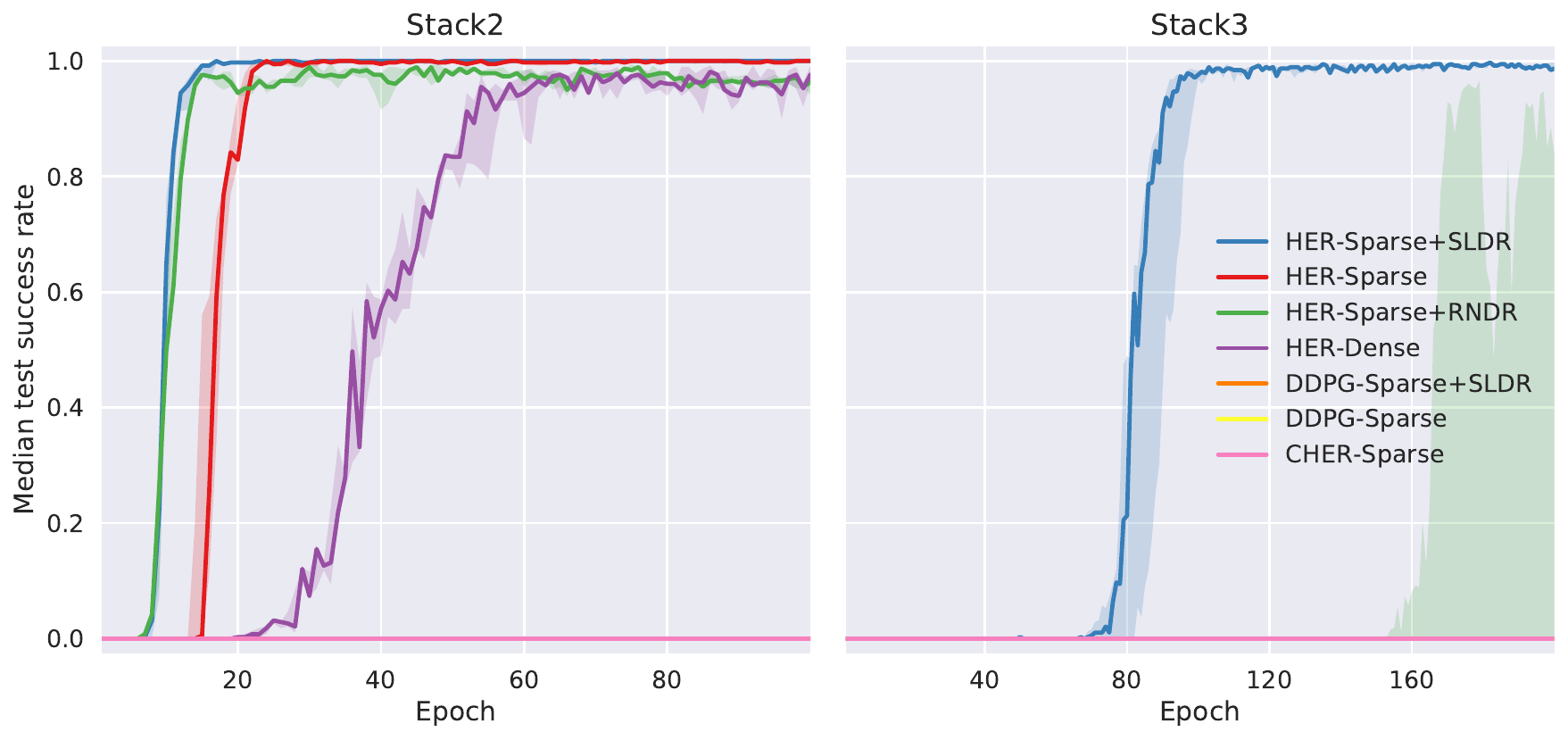}
    \caption{Fetch arm multi-object stacking.}
    \label{fig:learning_stack}
\end{subfigure}
\vskip 0.1in
\begin{subfigure}[b]{0.60\columnwidth}
    \centering
    \includegraphics[width=\columnwidth,trim={0.1cm 0.2cm 0.3cm 0.2cm},clip]{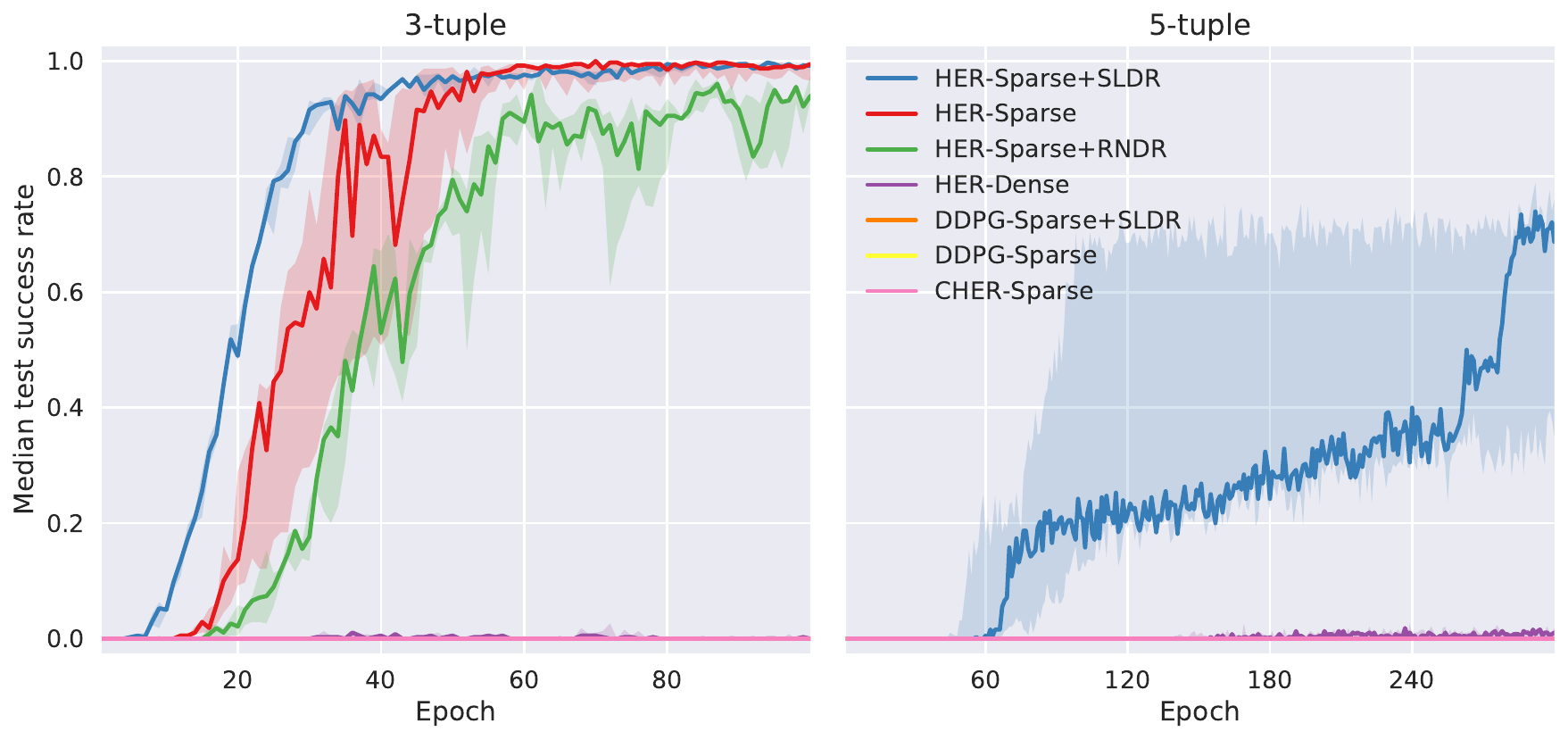}
    \caption{Fetch arm non-rigid object.}
    \label{fig:learning_flex}
\end{subfigure}
\caption{Learning curves of comparison algorithms on environments using Fetch robotic arm.}
\label{fig:learning_fetch_all}
\end{center}
\end{figure}

\begin{figure}
\begin{center}
\begin{subfigure}[b]{0.49\columnwidth}
    \centering
    \includegraphics[width=\columnwidth,trim={0.1cm 0.2cm 0.3cm 0.2cm},clip]{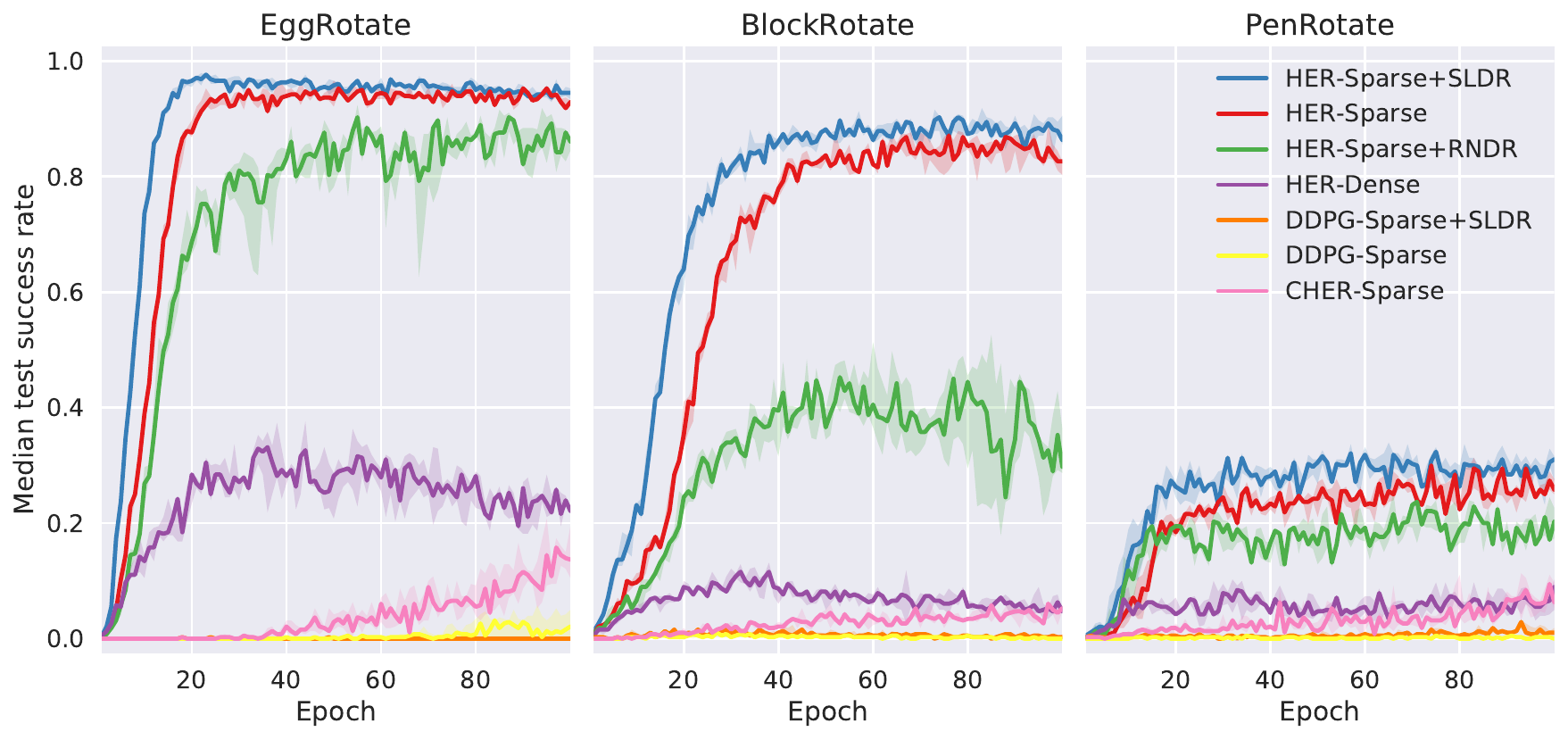}
    \caption{\textit{Rotate} variants.}
    \label{fig:learning_handrotate}
\end{subfigure}
\hfill
\begin{subfigure}[b]{0.49\columnwidth}
    \centering
    \includegraphics[width=\columnwidth,trim={0.1cm 0.2cm 0.3cm 0.2cm},clip]{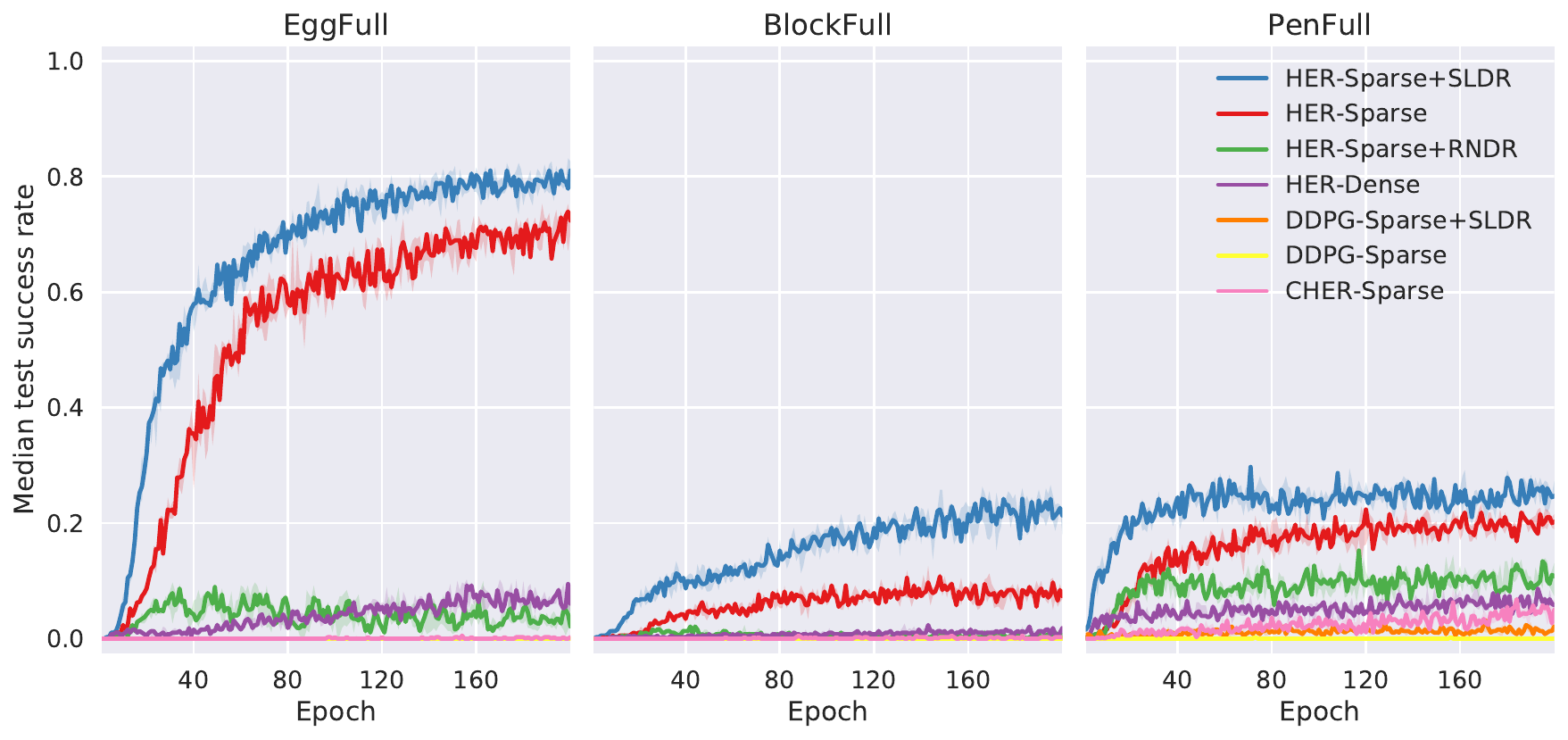}
    \caption{\textit{Full} variants.}
    \label{fig:learning_handfull}
\end{subfigure}
\caption{Learning curves of comparison algorithms on environments using Shadow robotic hand.}
\label{fig:learning_hand}
\end{center}
\end{figure}

The learning curves for Fetch, the \textit{Rotate} and \textit{Full} variants of Shadow's hand environments are reported in Figure \ref{fig:learning_fetch}, Figure \ref{fig:learning_handrotate} and Figure \ref{fig:learning_handfull}, respectively. We find that HER-Sparse+SLDR features a faster learning rate and the best performance on all the tasks. This evidence demonstrates that SLDR, coupled with DDPG and HER, can facilitate policy learning with sparse rewards. The benefits introduced by HER-Sparse+SLDR are particularly evident in hand manipulation tasks (Figure \ref{fig:learning_handrotate} and Figure \ref{fig:learning_handfull}) compared to fetch robot tasks (Figure \ref{fig:learning_fetch}), which are notoriously more complex to solve. Additionally, we find that HER-Sparse+SLDR outperforms HER-Sparse+RNDR in most tasks. A possible reason for this result is that most methods using auxiliary rewards are based on the notion of curiosity, whereby reaching unseen states is a preferable strategy, which is less suitable for manipulation tasks \cite{AndrychowiczCRS17}. In contrast, the proposed method exploits a notion of desired object locomotion to guide the main policy during training. We also observe that DDPG-Sparse+SLDR fails for most tasks. A possible reason for this is that, despite its effectiveness, the proposed approach still requires a suitable RL algorithm to learn from SLDR together with sparse environmental rewards. DDPG on its own is less effective for this task. We find that HER-Dense performs worse than HER-Sparse. This result support previous  observations that sparse rewards may be more beneficial for complex robot manipulation tasks compared to dense rewards \cite{AndrychowiczCRS17,PlappertHER2}. Finally, we observe that CHER-Sparse fails in most tasks and cannot facilitate successful learning. This is somewhat expected given our particular set up, and a possible explanation is in order. Sampling the replay buffer based on the proximity to true goals may work well for locomotion tasks because the distance between the robot gripper and the target is taken into account, and this distance is under direct control of the robot  from the very first episode. On the other hand, in the manipulation tasks, the distance between the object and the target stays roughly constant in the early training episodes as the robot has not yet learned to interact with the object. Such a sampling technique prioritising the replays depending on proximity may produce biased batches that can potentially disrupt the learning process. For example, a random robot action causing the object to move away from the target would favour trajectories characterised by a lack of interaction between the robot and the object. Although we report some success on \textit{EggRotate}, \textit{BlockRotate} and \textit{PenRotate} using CHER, this  is much lower than the success observed when using HER-Sparse+SLDR and HER-Sparse.

\subsection{Fetch Arm Multiple Object Environments} \label{sec:experiments_fetch_arm_multi_object}
For environments with $N$ objects, we reuse the locomotion policies trained on the \textit{PickAndPlace} task with single objects, and obtain an individual SLDR for each one of $N$ objects. We train $N+1$ action-value functions in total, i.e. one for each SLDR and one for the environmental sparse rewards. The manipulation policy is trained using the gradient in Eq. \ref{eq:policy_bot_multi_modify}. 

Inspired by \cite{PlappertHER2}, we randomly select between two initialisation settings for the training: (1) the targets are distributed on the table (i.e. an auxiliary task) and (2) the targets are stacked on top of each other (i.e. the original stacking task). Each initialisation setting is randomly selected with a probability of $0.5$. We have observed that this initialisation strategy helps HER-based methods complete the stacking tasks. From Figure \ref{fig:learning_stack}, we find that HER-Sparse+SLDR achieves better performance compared to HER-Sparse, HER-Sparse+RND and HER-Dense in the 2-object stacking task (\textit{Stack2}), while other methods fail. On the more complex 3-object stacking task (\textit{Stack3}), HER-Sparse+SLDR is the only algorithm to succeed. HER-Sparse+RND occasionally solves the \textit{Stack3} task with fixed random seeds but the performance is unstable across different random seeds and multiple runs. 
\begin{figure*}[hbt]
	\begin{center}
		\centerline{\includegraphics[width=\linewidth,trim={0.1cm 0.2cm 0.3cm 0.2cm},clip]{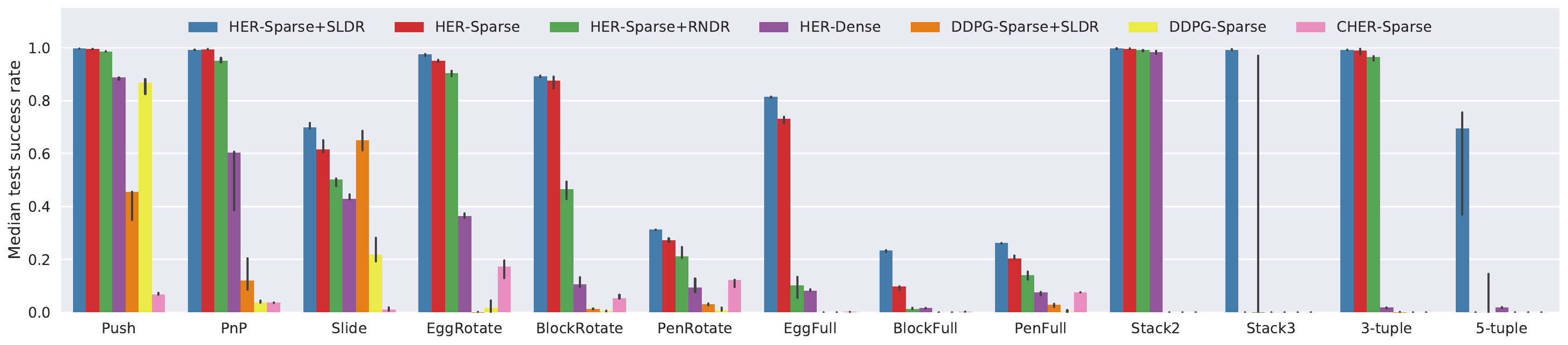}}
		\caption{Comparison of models with the best test success rate for all methods on all the environments.}
		\label{fig:best_model}
	\end{center}
\end{figure*}

\subsection{Fetch Arm Non-Rigid Object Environments}

The learning curves for 3-tuple and 5-tuple non-rigid object tasks are reported in Figure \ref{fig:learning_flex}. Similarly to the multiple object environment, HER-Sparse+SLDR achieves better performance for the 3-tuple task compared to HER-Sparse and HER-Sparse+RND, while the other methods fail to complete the task. For the more complex \textit{5-tuple} task, only HER-Sparse+SLDR is able to succeed. Among the 4 pre-defined targets depicted in Figure \ref{fig:flex_obj_task}, HER-Sparse+SLDR can achieve 3 targets on average, and can accomplish all 4 targets in one instance, out of 5 runs with different random seeds.

\subsection{Comparison Across the Best Models}

Figure \ref{fig:best_model} summarises the performance of the models with the best test success rates for each one of the competing methods. We can see that the proposed HER-Sparse+SLDR achieves top performance compared to all other methods. Specifically, HER-Sparse+SLDR is the only algorithm that is able to steadily solve 3-object stacking (\textit{Stack3}) and 5-tuple non-rigid object manipulation (\textit{5-tuple}). Remarkably, these two tasks have the highest complexity among all the 13 tasks. The \textit{Stack3} task includes multiple stages that require the robot to pick and place multiple objects with different source and target locations in a fixed order; in the \textit{5-tuple} task the object has the most complex dynamics. For these complex tasks, the proposed SLDR seems to be particularly beneficial. A possible reason is that, although the task is very complex, the objects are still able to learn good locomotion policies (see Fig~\ref{fig:learning_locomotion_fetch}) and the rewards learnt from locomotion policies provides critical feedback on how the object should be manipulated to complete the task. This type of object-based feedback is not utilised by other methods like HER and HER+RND. Our approach outperforms the runner-up by a large margin in the \textit{Full} variants of Shadow's hand manipulation tasks (\textit{EggFull}, \textit{BlockFull} and \textit{PenFull}), which feature complex state/action spaces and system dynamics. Finally, the proposed method consistently achieves better or similar performance than the runner-up in other simpler tasks. 

\section{Conclusion and Discussion} \label{sec:conclusion_discussion}

In this paper, we address the problem of mastering robot manipulation through deep reinforcement learning using only sparse rewards. The rationale for the proposed methodology is that robot manipulation tasks can be seen of as inducing object locomotion. Based on this observation, we propose to firstly  model the objects as independent entities that need to learn an optimal locomotion policy through interactions with a realistically simulated environment, then these policies are leveraged to improve the manipulation learning phase. 

\begin{figure}
\begin{center}
\begin{subfigure}[b]{0.49\columnwidth}
    \centering
    \includegraphics[width=\columnwidth,trim={0.1cm 0.2cm 0.3cm 0.2cm},clip]{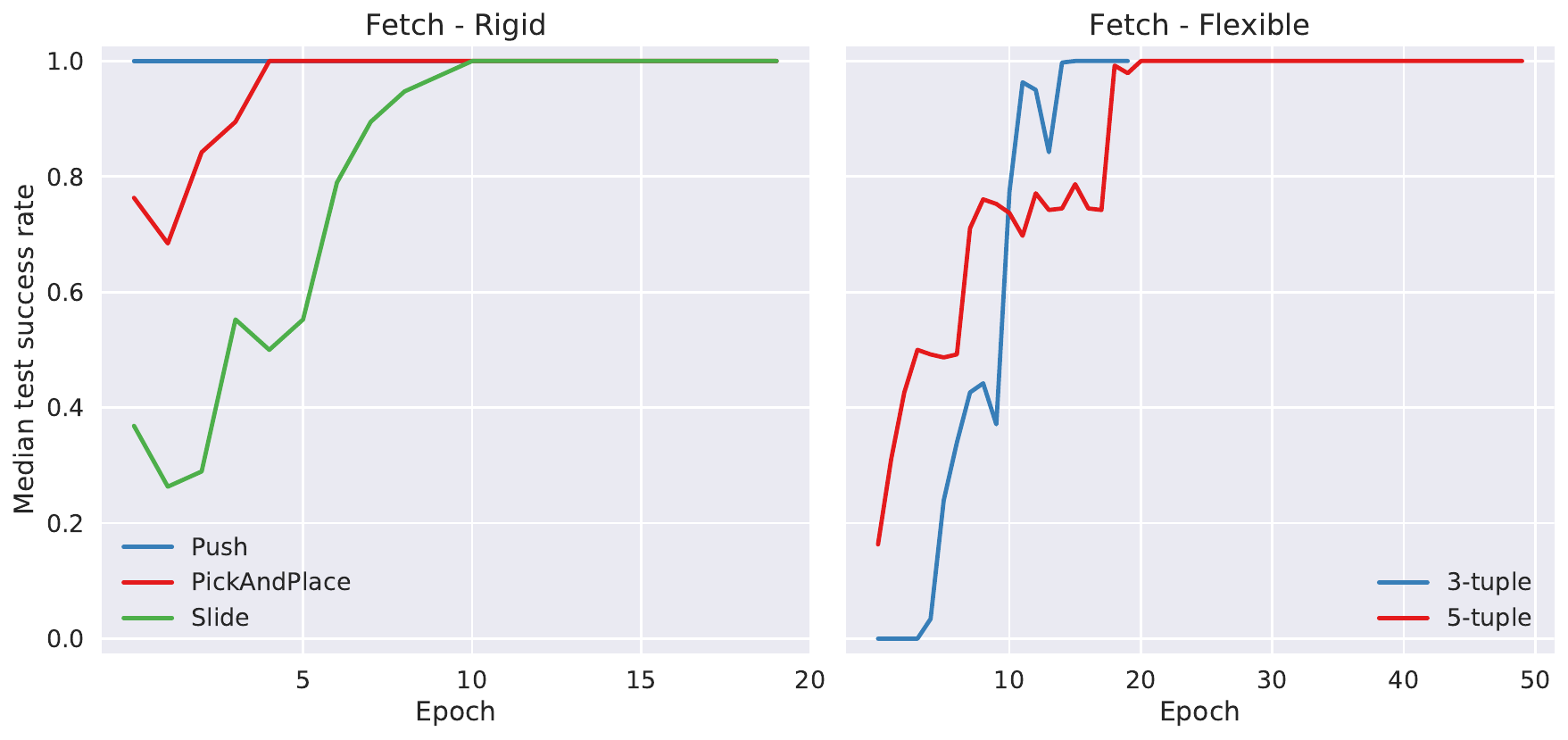}
    \caption{Fetch environments.}
    \label{fig:learning_locomotion_fetch}
\end{subfigure}
\hfill
\begin{subfigure}[b]{0.49\columnwidth}
    \centering
    \includegraphics[width=\columnwidth,trim={0.1cm 0.2cm 0.3cm 0.2cm},clip]{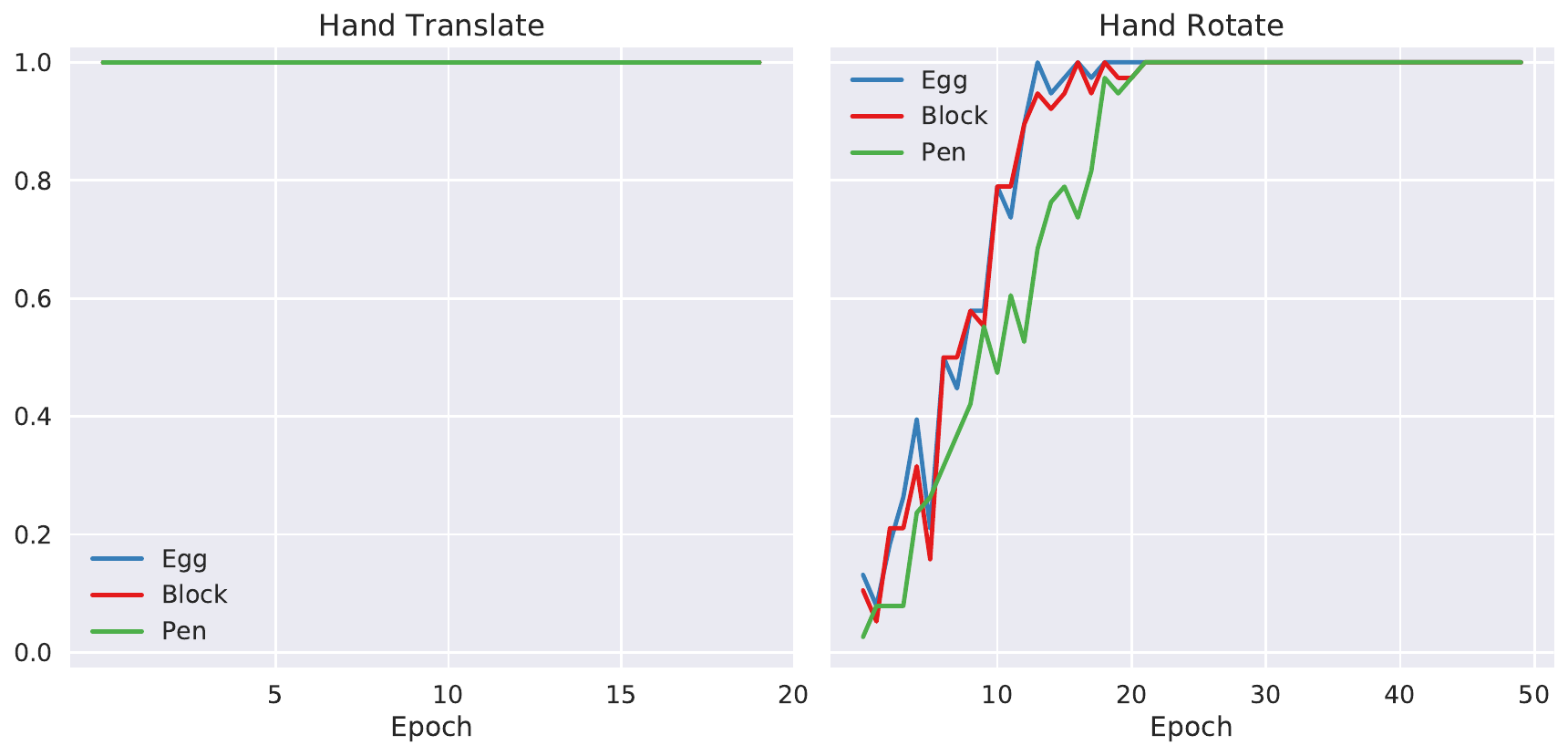}
    \caption{Shadow environments.}
    \label{fig:learning_locomotion_hand}
\end{subfigure}
\caption{Learning curves of for object locomotion policies.}
\label{fig:learning_locomotion}
\end{center}
\end{figure}

We believe that using SLDRs introduces significant advantages. First, SLDRs are generated artificially through a RL policy, hence require no human effort. Producing human demonstrations for complex tasks may prove difficult and/or costly to achieve without significant investments in human resources. For instance, it may be particularly difficult for a human to generate good demonstrations for tasks such as manipulating non-rigid objects  with a single hand or with a robotic gripper. On the other hand, we have demonstrated that the locomotion policies can be easily learnt, even for complex tasks, purely in a virtual environment; e.g., in our studies, these policies have achieved $100$\% success rate on all tasks (e.g. see Figure \ref{fig:learning_locomotion_fetch} and Figure \ref{fig:learning_locomotion_hand}). Furthermore, since the locomotion policy is learnt through RL, our proposed approach does not require task-specific domain knowledge and can be designed using only sparse rewards. Training the locomotion policies only requires the same sparse rewards provided by the environment hence the SLDRs produced through RL lead to high quality manipulation policies. This point has been supported by the empirical evidence obtained through experiments involving all 13 environments presented in this paper. As commonly observed in deep RL approaches, the use of neural networks as a function approximators for policies and inverse dynamics functions may introduce convergence issues and lead to non-optimal policies, but despite these limitations the proposed methodology has been proved to be sufficiently reliable and competitive. The proposed approach is orthogonal to existing methods that use expert demonstrations, and combining them together would be an interesting direction to be explored in the future. 

The performance of the proposed framework has been  thoroughly examined on 13 robot manipulation environments of increasing complexity. These studies demonstrate that faster learning and higher success rate can be achieved through SLDRs compared to existing methods. In our experiments, SLDRs have enabled the robots to solve complex tasks, such as stacking 3 objects and manipulating non-rigid object with 5 tuples, whereas competing methods have failed. Remarkably, we have been able to outperform runner-up methods by a significant margin for complex Shadow's hand manipulation tasks. Although SLDRs are obtained using a physics engine, this requirement does not restrict the applicability of the proposed approach to situations where the manipulation is learnt using real robot as long as the locomotion policy can  pre-learnt realistically. 

Several aspects will be investigated in follow-up work. We have noticed that when the interaction between the manipulating robot and the objects is very complex, the manipulation policy may be difficult to learn despite the fact that the locomotion policy is successfully learnt. For instance, in the case of the \textit{5-tuple} task with Fetch arm, although the locomotion policy achieves a $100$\% success rate (as shown in Figure \ref{fig:learning_locomotion_fetch}), the manipulation policy does not always completes the task (as shown in Figure \ref{fig:learning_flex} and Figure \ref{fig:best_model}). In such cases, when the ideal object locomotion depends heavily on the robot, the benefit of the SLDs is reduced. Another limitation is given by our Assumption 2 (Section~\ref{sec:robot_policy_learning}), which may not hold for some tasks. For example, for pen manipulation tasks with Shadow's hand, although the pen can rotate and translate itself to complete locomotion tasks (as shown in Figure \ref{fig:learning_locomotion_hand}), it is difficult for the robot to reproduce the same locomotion without dropping the pen. This issue can degrade the performance of the manipulation policy despite having obtained an optimal locomotion policy (see Figure \ref{fig:learning_handrotate}, Figure \ref{fig:learning_handfull} and Figure \ref{fig:best_model}). A possible solution would be to train the manipulation policy and locomotion policy jointly, and check whether the robot can reproduce the object locomotion suggested by the locomotion policy; a notion of ``reachability'' of object locomotion could be used to regularise the locomotion policy and enforce $P(z_t|\mu_\theta)\overset{d}{=}P(z_t|\nu_\theta)$. 

An important aspect to bear in mind is that our methodology requires the availability of a simulated environment for the application at hand. Nowadays, 
due to the well-documented sample inefficiency of most state-of-the-art, model-free DRL algorithms, such simulators are  commonly used for training RL policies before deployment in the real world. Besides, creating physically realistic environments from existing 3D models using modern tools has become almost effortless. In this sense, the  approach proposed in this work  requires only a marginal amount of additional engineering once a simulator has been developed. For instance, using MuJoCo, setting up the object locomotion policies would only entail the removal of the robot from the environment and inclusion of the objects as “mocap” entities. In comparison with other approaches, such as those relying on human demonstrations, the additional effort required to enable SLDR is only minimal.

In this paper we have adopted DDPG as the main training algorithm due to its widely reported effectiveness in continuous control tasks. However, our framework is sufficiently general, and other algorithms may be suitable such as trust region policy optimisation (TRPO)~\cite{schulman2015trust}, proximal policy optimisation (PPO)~\cite{PPO_Schulman_2017} and soft actor-critic~\cite{SoftActorCritic_Haarnoja_18}; analogously, model-based methods  ~\cite{ModelBasedRL_Chua_19,SOLAR_Zhang_19} could also provide feasible alternatives to be explored in future work.

\newpage
\bibliographystyle{unsrtnat}
\bibliography{bibliography}

\end{document}